\documentclass{article}
\usepackage{imav}
\usepackage{times}
\usepackage{graphicx}
\usepackage{amsmath}
\usepackage{makecell}
\usepackage{amssymb}
\usepackage{booktabs}
\usepackage[nointegrals]{wasysym}
\usepackage{cite}
\usepackage[bookmarks=true,colorlinks=true]{hyperref}
\usepackage{tabularx} 

\title{VLN on the Fly: An Onboard Vision-Language Navigation Stack for Aerial Robots}

\author{Marco S. Tayar\textsuperscript{\textdagger, *}, Felipe Tommaselli\textsuperscript{*}, Gianluca Capezzuto\textsuperscript{*}, Pedro Antonio Rabelo Saraiva\textsuperscript{*}, \\
Pedro H. V. de Freitas, Lucas Kido, Guilherme Sonego, Ricardo V. Godoy, and Marcelo Becker \\
University of S\~ao Paulo (USP), Brazil \\[4pt]
{\small\textsuperscript{*}These authors contributed equally to this work.}
{\small\textsuperscript{\textdagger} Corresponding author: marcotayar@usp.br}}
\begin{document}

\maketitle
\thispagestyle{empty} 

\begin{abstract}
Running vision-language navigation fully onboard an aerial robot is hard, since grounding, planning, and control must share limited compute and a single-stage error is difficult to isolate in flight. End-to-end aerial policies fuse these stages into one network, giving up the observability and safety checks a modular stack keeps available. We propose VLN on the Fly, an onboard stack that keeps grounding, planning, and control as separate, inspectable stages. A quantized VLM grounds an instruction to a coarse image cell, depth lifts it to a 3D goal, a fast B-spline planner returns a feasible trajectory, and a pretrained reinforcement learning policy tracks it to motor commands across quadrotors. Across 15 onboard flights over three everyday referents in a controlled indoor volume, the stack reaches the target in 13 of 15 trials with 5.72 cm mean goal error and 39.3\% average GPU utilization. In 6 additional cluttered-environment trials, the stack tracks collision-free trajectories under onboard perception gating. Project page: \href{https://vln-on-the-fly.github.io}{\nolinkurl{vln-on-the-fly.github.io}}
\end{abstract}

\section{Introduction}

\begin{figure}[htbp]
    \centering
    \includegraphics[width=\linewidth]{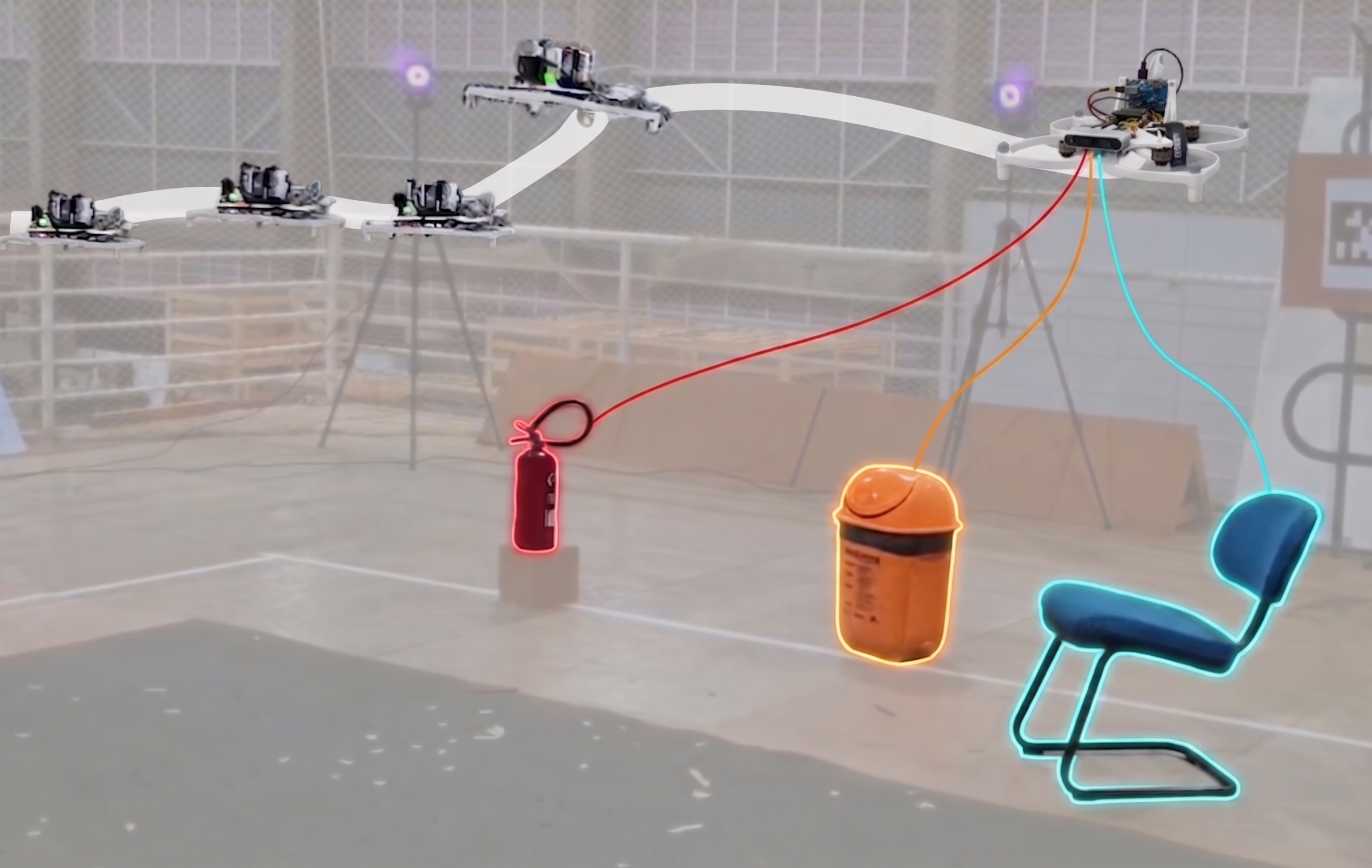}
    \caption{\textbf{VLN on the Fly}, an onboard vision-language navigation stack that grounds language instructions and plans and executes trajectories toward named referents, here a fire extinguisher, a trash bin, or a chair.}
    \label{fig:first_page}
\end{figure}

Aerial robots are increasingly being deployed in human-centered indoor environments, including warehouses, offices, inspection sites, and residential spaces. Indoor deployment introduces GPS denial, cluttered geometry, narrow passages, and layouts built around people or ground vehicles rather than autonomous flight. Operation under such conditions depends on the platform's robustness and the ability of the underlying task-completion models to generalize beyond a fixed environment. Language-conditioned control is promising in this setting because natural-language instructions allow a human to specify task intent directly in a new scenario, without re-engineering the system for each task.

Vision-Language-Action (VLA) models have been used to address this need, with a single end-to-end policy trained to map raw observations and instructions directly to control commands. Applied to navigation, this class is commonly referred to as Vision-Language Navigation (VLN)~\cite{zhou2025beast, reuss2025flower, shah2023gnm, navid4d2025}, and recent aerial works such as AerialVLA~\cite{AerialVLA2026}, AutoFly~\cite{AutoFly2026}, and Singer~\cite{Adang2025Singer} point in promising directions. However, the compositional generalization behind the success of language models has proven harder to achieve in physical autonomy~\cite{zhang2024vision}. For aerial robots, each new instruction must be grounded not only in semantic context but also in feasible motion under vehicle dynamics and safety constraints. When grounding, spatial reasoning, and control are encoded inside a single end-to-end policy, limited observability makes failures harder to localize and safety constraints harder to inspect or enforce through standard planning mechanisms such as collision checks, feasibility tests, and recovery behaviors~\cite{Kim2026Guardrails}.

Recent literature shows that decoupling these models enables desirable behaviors in navigation stacks~\cite{werby2024hierarchical, wei2025dualvln}, since separate modules remain observable, reusable path-planning safety guards remain available, and changes to one component of the stack do not require retraining the entire policy. For a decoupled approach, integrating the hierarchical stack, particularly with the flight controller's inner modules, remains a valid concern underexplored in the literature. Concretely, we address three concerns: (i) safety, (ii) hierarchical complexity, and (iii) onboard compute, since running a local Vision-Language Model (VLM) together with a full stack onboard is inherently challenging.

In this work, we propose \textbf{VLN on the Fly}, an onboard stack for navigation on aerial vehicles (Fig.~\ref{fig:first_page}). We embed a quantized VLM (Qwen-3.5-2B~\cite{qwen3.5}) for coarse grid grounding, where the model identifies the image region containing the target object from the language instruction and visual scene. The selected region is combined with depth information to estimate a 3D goal, which is passed to EGO-Planner~\cite{zhou2020egoplanneresdffreegradientbasedlocal} for B-spline trajectory generation. To overcome the low-level complexity of the hierarchical pipeline, we leverage the pretrained RAPTOR policy~\cite{eschmann2026raptorfoundationpolicyquadrotor}, which maps a position setpoint and the proprioceptive data directly to motor outputs. To maintain safety across this hierarchy, we validate the projected goal against the operating bounds before planning, while the planner's occupancy map supports collision-aware trajectory generation. We investigate these questions through system design, real-flight evaluation, and per-module characterization. Our contributions are:
\begin{itemize}
    \item An onboard decoupled aerial VLN stack combining a quantized VLM for open-vocabulary grounding, EGO-Planner for 3D B-spline planning, and pretrained RAPTOR for low-level control, without end-to-end training or offboard compute.
    \item A lightweight safety supervisor, using a finite-state machine to reject projected goals outside the operating bounds and to gate trajectory execution based on planner feasibility from the onboard occupancy map.
    \item A real-flight feasibility evaluation of the full onboard stack on an open-vocabulary object-goal task over three referents, and a per-stage failure attribution enabled by the modular design.
\end{itemize}

\section{Related Work}

\textbf{Aerial VLN} has largely been end-to-end, a single network mapping instructions and egocentric observations to actions, from the AerialVLN benchmark~\cite{liu2023aerialvln} to recent open-vocabulary models~\cite{AutoFly2026, Adang2025Singer}, with VLA-AN~\cite{wu2025vlaan} carrying such a policy fully onboard. This coupling forfeits the observability and path-planning safety guards of a conventional stack and ties the policy to one platform. In contrast, VLN on the Fly keeps language grounding, metric planning, and low-level control as separate stages, enabling onboard VLN while preserving inspectable interfaces.

\textbf{Decoupled VLNs} split semantic reasoning from control to recover those guards. DualVLN~\cite{wei2025dualvln} separates high and low-level execution, while See-Point-Fly~\cite{hu2025spf} and Fly0~\cite{xu2026fly0} ground an instruction to an image point, the latter projects it to a 3D target with depth for a geometric planner, the same mid-stack we adopt. AirHunt~\cite{chen2026airhunt} keeps the slow VLM from bottlenecking the planner, and OnFly~\cite{zheng2026onfly} runs the VLM fully onboard on our class of embedded computers, confirming the recipe is both effective and onboard-capable. We adopt the same decoupling principle, but replace pixel-level grounding with coarse-grid grounding and connect the VLM-to-planner interface to a learned flight policy.

\textbf{Reinforcement Learning Policies} map vehicle state directly to actuator commands at the high control rates required for agile flight, around 100\,Hz, as shown in champion-level drone racing~\cite{kaufmann2023champion} or autonomous inspection~\cite{LARS}. However, many learned controllers specialize in a single airframe and require renewed system identification or retraining after hardware changes. RAPTOR~\cite{eschmann2026raptorfoundationpolicyquadrotor} addresses this issue with a foundation policy that adapts across quadrotors zero-shot. We use RAPTOR as the low-level stage of the decoupled VLN stack, feeding position setpoints directly from the planner and avoiding a hand-tuned position-attitude cascade.

\begin{figure*}[t!]
    \centering
    \includegraphics[width=0.8\linewidth]{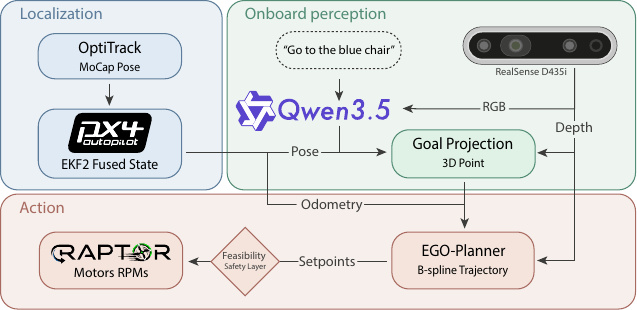}
    \caption{\textbf{Overview of the VLN on the Fly.} The RGB-D camera and language instruction are processed onboard by Qwen3.5 to project a 3D goal, EGO-Planner generates a feasible B-spline trajectory, and RAPTOR tracks the resulting setpoints as motor commands. OptiTrack pose is fused by EKF2 for localization in the experimental setup, while the safety layer gates motion.}
    \label{fig:overview}
\end{figure*}

\section{System Design} \label{sec:system}

Our framework presents a modular, hierarchical, and fully onboard VLN architecture with three processing stages coordinated by a safety supervisor (Fig.~\ref{fig:overview}). The grounding stage converts the instruction and current observation into a 3D goal, EGO-Planner generates a feasible trajectory, and RAPTOR tracks the resulting setpoints. Explicit goals and setpoints keep the interfaces observable, while the supervisor validates candidate goals and gates trajectory execution.

\subsection{VLM-based Goal Grounding}
\label{subsec:vlm}

We use Qwen-3.5-2B quantized to INT4~\cite{qwen3.5}, reducing memory use enough for onboard inference. At each VLM query, the model takes the egocentric RGB frame and the instruction, then returns one of nine named regions from a $3\times3$ image grid together with a confidence score, or reports that the target is not visible. The grid is defined only in the text prompt and is not rendered onto the image. The returned region is mapped to the corresponding cell during post-processing. Restricting the output to nine region names removes pixel coordinates, making the result easier to check.

For the returned cell, the median valid depth $d_{\mathrm{cell}}$ is assigned to its center pixel $(u_{\mathrm{cell}},v_{\mathrm{cell}})$. The corresponding point in the camera frame is obtained through the back-projection in~\eqref{eq:goal_projection}:
\begin{equation}
    \mathbf{p}_{\mathrm{camera}}
    =
    d_{\mathrm{cell}}
    \begin{bmatrix}
        (u_{\mathrm{cell}}-c_x)/f_x \\
        (v_{\mathrm{cell}}-c_y)/f_y \\
        1
    \end{bmatrix},
    \label{eq:goal_projection}
\end{equation}
where $f_x$ and $f_y$ are the camera focal lengths, and $(c_x,c_y)$ is the principal point. The current pose then transforms the point into the map frame to obtain the planner goal. Using the cell median reduces sensitivity to isolated invalid or noisy depth readings. If the cell contains no valid depth within the sensor range, the goal is rejected.

The goal altitude is constrained to a narrow band around the current flight altitude, keeping the goal within a safe vertical range and avoiding climbs or descents caused by depth noise. Between VLM queries, the planner and policy continue running at their own rates.

\subsection{Safety Supervisor}
\label{subsec:fsm}

The safety supervisor controls the handoff between VLM grounding and EGO-Planner. At take-off, the initial vehicle position defines the center of a bounded operating volume. A candidate goal $(x_g,y_g,z_g)$ is accepted only when
\begin{equation}
    |x_g-x_0| \leq b_x,\qquad
    |y_g-y_0| \leq b_y,\qquad
    |z_g-z_0| \leq b_z,
    \label{eq:operating_bounds}
\end{equation}
where $(x_0,y_0,z_0)$ denotes the initial vehicle position and $b_x$, $b_y$, and $b_z$ define the configured limits along each axis. Goals violating~\eqref{eq:operating_bounds} are rejected before planning.

Before goal publication, the supervisor requires three consecutive predictions for the same grid region, each above the configured confidence threshold. Only predictions with a valid projected goal enter the consensus process. After acceptance, one goal remains active until EGO-Planner reports completion, rejection, or failure. Nearby proposals and rapid updates are suppressed to prevent unstable commands between VLM queries.

\subsection{3D Planning}
\label{subsec:planner}

The validated 3D goal is passed to EGO-Planner~\cite{zhou2020egoplanneresdffreegradientbasedlocal}, which returns a smooth, dynamically feasible trajectory. The planner does not require a Euclidean Signed Distance Field (ESDF), allowing local, on-demand collision evaluation without constructing a full distance field. The lower computational cost allows EGO-Planner to run alongside the VLM on the same onboard computer. EGO-Planner represents trajectories as uniform B-splines and plans natively in 3D.

The B-spline convex-hull property allows velocity, acceleration, and collision constraints to be applied through the control points. The corresponding limits are configured for each airframe, defining how the platform's motion limits are passed to the planner. Local support also confines each control-point update to a short segment of the trajectory.

After goal validation, EGO-Planner checks the trajectory's feasibility against a local occupancy map built from the depth camera and the current pose. The map is entered into the collision cost before the optimized B-spline is sampled into position setpoints. Replanning refreshes the setpoints.

\subsection{Low-level Policy}
\label{subsec:policy}

A conventional flight-control stack maps planner setpoints to motor commands through position, velocity, attitude, and rate controllers, each tuned to the drone's mass, inertia, motor constants, and thrust limits. RAPTOR~\cite{eschmann2026raptorfoundationpolicyquadrotor} replaces the cascade with a pretrained policy that maps the setpoint-relative state directly to actuator commands.

RAPTOR receives the position setpoints from Section~\ref{subsec:planner} relative to the current vehicle state. A lightweight tracking bridge samples the planned path as a moving reference and provides position setpoints with velocity feedforward. After replanning, the reference resumes from the path point nearest the current vehicle position, reducing discontinuities between successive trajectories. Updating the stream after replanning ensures the policy tracks the latest trajectory. RAPTOR handles low-level control through a network with 2,084 parameters and three recurrent layers, running directly on the flight controller without retraining for our airframe.

\begin{figure}[t]
    \centering
    \frame{\includegraphics[width=0.85\linewidth]{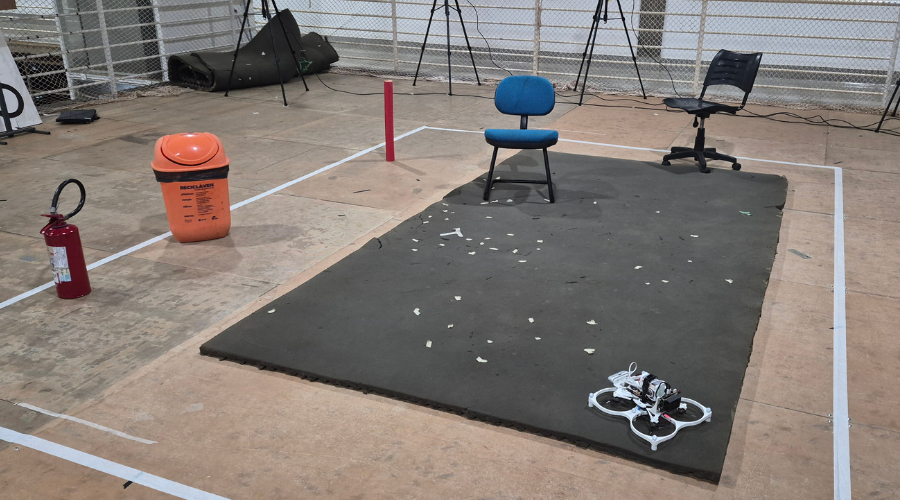}}\\[5pt]
    \frame{\includegraphics[width=0.85\linewidth]{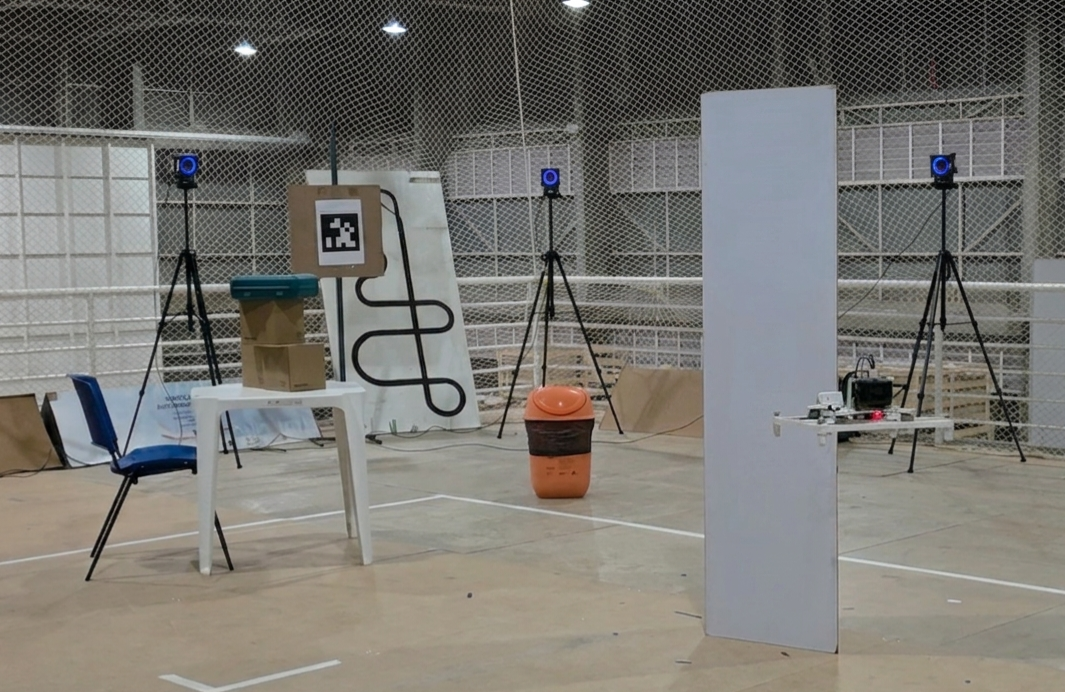}}
    \caption{\textbf{Experimental Setup.} \textbf{(a)} Controlled indoor flight volume for the open-vocabulary object-goal trials, with the quadrotor, marked operating bounds, and everyday referents. \textbf{(b)} Cluttered setup for the path-planning trials (Sec.~\ref{subsec:obstacles}), with a table, chair, tool case, and a freestanding vertical panel placed between the take-off pose and the trash-bin referent.}
    \label{fig:xp_scenario}
    \vspace{-0.1in}
\end{figure}

\section{Experimental Results}
\label{sec:experiments}

In this section, we evaluate VLN on the Fly as an onboard aerial VLN stack for open-vocabulary object-goal navigation. We first describe the platform and experimental setup, then evaluate goal-reaching performance across 15 real flights and profile onboard VLM latency. We next analyze the perception design through grid-resolution, depth-readout, and referent-discrimination studies. Finally, we test the complete stack in cluttered scenes to assess goal validation, obstacle-aware planning, and trajectory tracking.

We deploy VLN on the Fly on MIRA~\cite{deoliveira2026mira}, an open-source quadrotor equipped with an Intel RealSense D435i RGB-D camera, an onboard Jetson Orin NX, and a Pixhawk 6C flight controller. An OptiTrack PrimeX 41 motion-capture system provides pose estimates used as odometry.

\subsection{Performance Analysis}

Trials are conducted as shown in Fig.~\ref{fig:xp_scenario}, with three everyday referents: a trash bin, a chair, and a fire extinguisher. For each referent, we run the full stack 5 times, for a total of 15 onboard flights from the same take-off pose. Each flight receives a single instruction naming the target. The scene layout remains fixed during each run.

\begin{figure*}[t]
    \centering
    \frame{\includegraphics[width=0.8\textwidth]{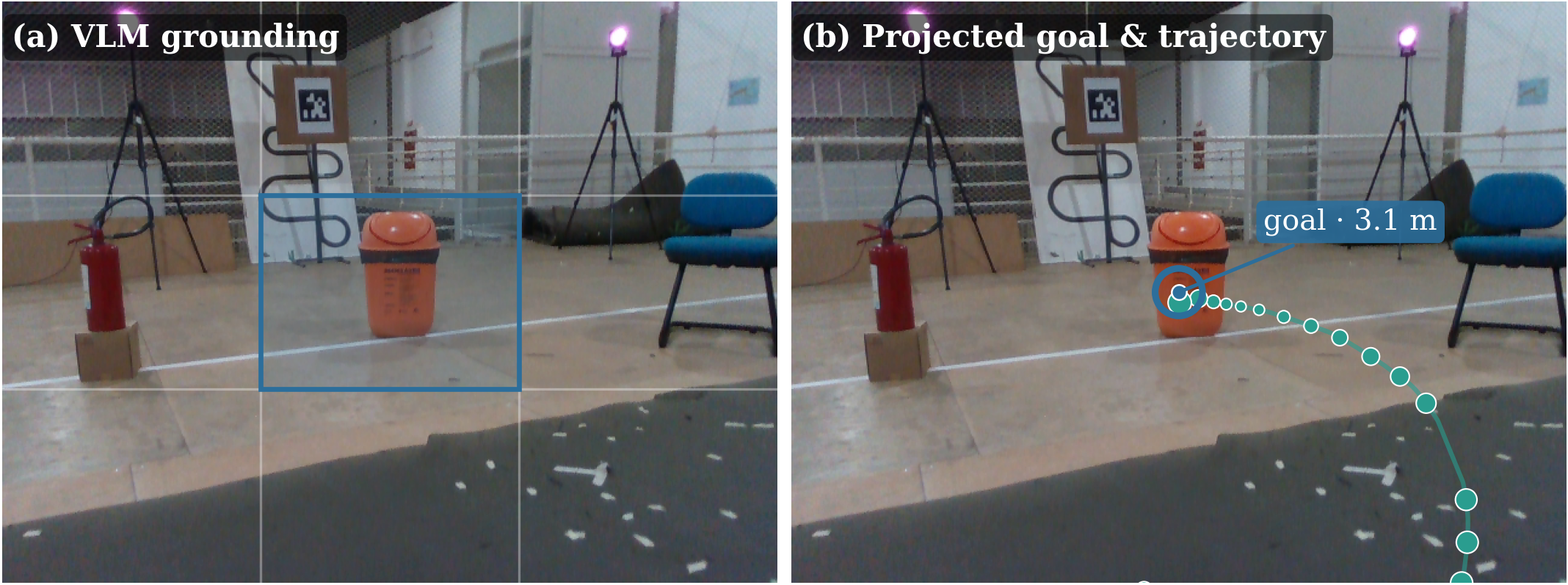}}
    \caption{\textbf{Qualitative Grounding and Planning Result.} A single onboard frame from a representative successful run for the instruction \emph{``go to the trash bin''}, with the fire extinguisher and chair present as distractors. \textbf{(a)} VLM grounding over the image grid: the selected center cell (blue) localizes the referent. \textbf{(b)} The same frame as a navigation command: the back-projected 3D goal at the selected cell's ray, and the executed onboard trajectory projected into the camera view.}
    \label{fig:qualitative}
    \vspace{-0.1in}
\end{figure*}

As reported in Table~\ref{tab:results}, our method reached the commanded referent in 13 of 15 flights (87\% success rate, with a Wilson 95\% confidence interval of 0.62--0.96). The trash bin succeeded in all five runs, ending directly above the referent every time (zero goal error) even though the executed paths curved on the way in. The chair and fire extinguisher each failed once, and in both cases the vehicle approached the correct object but stopped short of the 20 cm radius once the target left the camera view, preventing goal refinement.

The qualitative run in Fig.~\ref{fig:qualitative} exposes the main interfaces of the stack during a successful run. The VLM output is correct at the coarse-region level, supporting our choice to avoid pixel-level grounding. However, the main residual error appears downstream, as the limited RGB-D range and local inflation in the occupancy map can bias the planner away from a straight approach, as seen in the curved projected trajectory.

\begin{table}[t]
    \centering
    \caption{Open-Vocabulary Object-Goal Navigation}
    \label{tab:results}
    \resizebox{\linewidth}{!}{
    \begin{tabular}{lcccc}
    \toprule
    \textbf{Referent} & \textbf{\makecell{Success \\ rate\,$\uparrow$}} & \textbf{\makecell{Goal err. \\ (cm)\,$\downarrow$}} & \textbf{\makecell{Track err. \\ (cm)\,$\downarrow$}} & \textbf{\makecell{Avg. GPU \\ (\%)}} \\
    \midrule
    Trash bin & 1.00 & 0.00 & 34.96 & 42.2 \\
    Chair & 0.80 & 10.07 & 26.13 & 37.3 \\
    Fire extinguisher & 0.80 & 7.08 & 37.64 & 38.4 \\
    \midrule
    \textbf{Overall} & \textbf{0.87} & \textbf{5.72} & \textbf{32.91} & \textbf{39.3} \\
    \bottomrule
    \end{tabular}
    }
\end{table}

Beyond the GPU utilization of Table~\ref{tab:results}, we profile the grounding stage directly from the onboard inference records (Table~\ref{tab:latency}, $N{=}1865$ queries). On the Jetson Orin NX the quantized 2B VLM completes a grounding query in a median of 0.79\,s (95th percentile 0.88\,s), which comfortably sustains the 0.5\,Hz query rate and leaves the planner and policy to run uninterrupted in between. Time-to-first-token, which covers vision encoding and prompt prefill, dominates and is nearly constant at 0.60\,s, the total time varies only with the short region output, so latency is stable across queries. 

\begin{table}[h]
    \centering
    \caption{Onboard VLM Inference Timing}
    \label{tab:latency}
    \begin{tabular}{lccc}
    \toprule
    \textbf{Stage} & \textbf{Median} & \textbf{95th pct.} & \textbf{Max} \\
    \midrule
    Time-to-first-token & 0.60\,s & 0.66\,s & 0.72\,s \\
    Total inference     & 0.79\,s & 0.88\,s & 1.12\,s \\
    \bottomrule
    \end{tabular}
\end{table}
\subsection{Perception Ablation Study}
\label{subsec:profiling}

To isolate the effect of otherwise untracked hyper-parameters and to motivate our design choice, Table~\ref{tab:grid} presents an offline analysis of the perception layer. We replayed the logged onboard depth frames using the same cell-selection and projection pipeline as in flight. In addition to the valid-goal rate against the ground-truth position, we report the inter-frame goal-selection jitter. 

A $1\times1$ grid provides stable depth estimates but no image-plane localization and therefore cannot spatially distinguish multiple referents. Among grids that preserve spatial localization, $3\times3$ achieves a valid-goal rate of 1.00 with substantially lower jitter than $5\times5$, motivating its adoption as our default configuration.

\begin{table}[!t]
    \centering
    \caption{Grounding Resolution Ablation}
    \label{tab:grid}
    \resizebox{\linewidth}{!}{
    \begin{tabular}{lcc}
    \toprule
    \textbf{Region grid} & \textbf{Valid-goal rate}\,$\uparrow$ & \textbf{Selection jitter (m)}\,$\downarrow$ \\
    \midrule
    $1\times1$        & 0.94 & 0.07 \\
    $3\times3$ (ours) & 1.00 & 0.70 \\
    $5\times5$        & 1.00 & 1.27 \\
    Single pixel      & 0.15 & 1.16 \\
    \bottomrule
    \end{tabular}
    }
\end{table}

Taking the grid to its limit, a single pixel, removes the region abstraction and makes explicit why we ground at the cell level (bottom of Table~\ref{tab:grid}). Back-projecting the goal from one image pixel yields a valid depth reading in only $\sim$15\% of frames (invalid in ~85\%), since a lone pixel frequently lands on a low-texture, reflective, or out-of-range surface, and where it is valid, it can read across an object boundary and back-project from the background behind the referent, differing from the cell median by up to 0.8\,m. The selected $3\times3$ cell median avoids these failures.

To further support the open-vocabulary claim, we replayed identical onboard frames through the VLM, varying only the referent name in the prompt ($N{=}1865$ queries, Table~\ref{tab:grounding}). On frames where the model commits to a region, the two present referents map to disjoint, scene-consistent locations, the chair to the top-left (86\%) and the trash bin to the center (90\%). The absent referent is correctly reported as not visible in 66\% of frames, but when it does commit, it defaults to the center (79\%), the trash bin's region. This hallucination mode does not affect the discrimination between present referents, but it suggests trusting not-visible reports over commitments to out-of-scene targets. Although the VLM exhibits a high false-negative rate for present objects (44--53\% not visible), the tracking and projection pipeline successfully bridges these intermittent perception gaps.

\begin{table}[t]
    \centering
    \caption{VLM Region Selection by Prompted Referent}
    \label{tab:grounding}
    \small
    \setlength{\tabcolsep}{4pt}
    \begin{tabular}{lccc}
    \toprule
    \textbf{Prompted referent} & \textbf{Not visible} & \textbf{Modal region} & \textbf{Share} \\
    \midrule
    Chair (left)               & 53\% & Top-left & 86\% \\
    Trash bin (center)         & 44\% & Center   & 90\% \\
    Fire extinguisher (absent) & 66\% & Center   & 79\% \\
    \bottomrule
    \end{tabular}
\end{table}

\subsection{Cluttered-Environment Evaluation}
\label{subsec:obstacles}
\begin{figure}[t]
    \centering
    \includegraphics[width=0.92\linewidth]{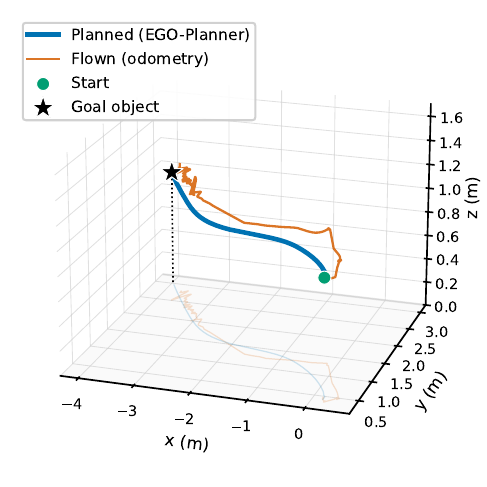}
    \caption{\textbf{Planned vs.\ Flown Trajectory in Clutter.} EGO-Planner's collision-free B-spline (blue) runs from take-off (green) to the goal object (star), and RAPTOR tracks it (orange). Altitude is held near the commanded band, faint floor shadows show the horizontal path. The window spans trajectory generation to the first goal-reach.}
    \label{fig:obstacle_traj}
\end{figure}

The trials above isolate grounding accuracy in a largely open volume. To probe the stack under the cluttered conditions typical of indoor deployment, we ran a small set of additional onboard flights ($n=6$) toward the trash-bin referent with obstacles (a chair, a tool case, and mocap tripods) placed between the take-off pose and the target. Using the onboard occupancy map, EGO-Planner produced collision-free B-spline trajectories that RAPTOR tracked to the referent. In three of the six runs the stack reached the referent collision-free, ending 0.00--0.183\,m from the goal. In the other two the safety layer never released a goal that satisfied its depth and bounds checks, so no trajectory was executed and no collision occurred. In the remaining run (Run 3), the vehicle successfully avoided obstacles but terminated just outside the success radius (0.21\,m from the goal) due to high tracking deviation. This demonstrates that the same onboard stack extends from open-volume goal reaching to cluttered navigation without changes, and that the goal-validation interface degrades to a safe no-commit. Fig.~\ref{fig:obstacle_traj} shows the planned and flown trajectories for a representative run, and Table~\ref{tab:obstacle} reports per-run errors measured from trajectory generation to the first goal-reach. Across the four executed runs the flown path stayed close to the planned B-spline, with a mean tracking error of $17.6$\,cm (per-run means $10.9$--$29.9$\,cm), altitude was held near the commanded band, so the horizontal path carried the navigation while the vehicle navigated the clutter.

\section{Limitations}
As with many open-vocabulary systems, the generalization of our results is not fully characterized. While experiments demonstrate the feasibility of the proposed VLN tasks, gains in modularity and safety come at the cost of integration trade-offs. In particular, the target remains fixed between goal queries, so a grounding error can persist until the next query, an asynchronous design could refresh it sooner. Grounding also operates on the current egocentric frame alone, with no spatial memory, so a referent outside the field of view must first be brought into view before it can be grounded, which bounds the contextual and sequential task families. Finally, depth back-projection proved to be limited in practice by the stereo sensor's range and by invalid depth regions.

To isolate the present contributions, we made two simplifying assumptions that future work could relax: fixed-altitude flight and external odometry. The $z$-clamp fixes the altitude per environment, so the planner's full 3D capability is not exercised. Moreover, all trials are conducted within a single OptiTrack volume from a known initial pose, without evaluating arbitrary initialization or relocalization.

\begin{table}[!t]
    \centering
    \caption{Goal and Tracking Errors for Cluttered Flights}
    \label{tab:obstacle}
    \resizebox{\linewidth}{!}{
    \begin{tabular}{lccc}
    \toprule
    \textbf{Run} & \textbf{\makecell{Goal err. \\ (cm)\,$\downarrow$}} & \textbf{\makecell{Mean track. err. \\ (cm)\,$\downarrow$}} & \textbf{\makecell{p90 track. err. \\ (cm)\,$\downarrow$}} \\
    \midrule
    1 & 0.00  & 14.5 & 18.6 \\
    2 & 18.3 & 15.1 & 26.4 \\
    3 & 20.7 & 29.9 & 79.1 \\
    4 & 0.00 & 10.9 & 21.2 \\
    \midrule
    \textbf{Mean} & \textbf{9.75} & \textbf{17.6} & \textbf{36.3} \\
    \bottomrule
    \end{tabular}
    }
\end{table}

\section{Conclusion}

We presented \textbf{VLN on the Fly}, an onboard decoupled aerial VLN stack that grounds natural-language instructions into 3D goals with a quantized VLM, plans collision-aware trajectories with EGO-Planner~\cite{zhou2020egoplanneresdffreegradientbasedlocal}, and executes position setpoints through the pretrained RAPTOR policy~\cite{eschmann2026raptorfoundationpolicyquadrotor}. The decoupled design preserves intermediate observability, allowing grounding, goal projection, planning, and control errors to be inspected separately during real flight. Our experiments characterize open-vocabulary object-goal navigation in a controlled indoor volume, demonstrating onboard execution while highlighting remaining sensitivity to depth range, metric projection, and planner-interface effects. Complementary sensing, longer-range depth, and tighter coupling between perception uncertainty and trajectory generation could improve robustness in less constrained environments and provide a clearer path toward longer-horizon aerial VLN.

\section{Acknowledgments}

This work was supported in part by the S\~ao Paulo Research Foundation (FAPESP), Grants \#2025/20858-7 and \#2025/22381-3; in part by the Conselho Nacional de Desenvolvimento Cient\'ifico e Tecnol\'ogico (CNPq), Grant 308092/2020-1; and by Petr\'{o}leo Brasileiro S/A - Petrobras, using resources from the ANP R\&D clause, in partnership with the University of S\~{a}o Paulo~(USP) and the Funda\c{c}\~{a}o de Apoio \`{a} F\'{\i}sica e \`{a} Qu\'{\i}mica~(FAFQ), under Cooperation Agreements \#2023/00016-6 and \#2023/00013-7.

\bibliographystyle{unsrt}
\bibliography{imav_bibliography}

@inproceedings{zhou2025beast,
 author = {Zhou, Hongyi and Liao, Weiran and Huang, Xi and Tang, Yucheng and Otto, Fabian and Jia, Xiaogang and Jiang, Xinkai and Hilber, Simon and Li, Ge and Wang, Qian and Ya\u{g}murlu, \"{O}mer and Blank, Nils and Reuss, Moritz and Lioutikov, Rudolf},
 booktitle = {Advances in Neural Information Processing Systems},
 pages = {172934--172959},
 title = {BEAST: Efficient Tokenization of B-Splines Encoded Action Sequences for Imitation Learning},
 volume = {38},
 year = {2025}
}

@misc{reuss2025flower,
  title={FLOWER: Democratizing Generalist Robot Policies with Efficient Vision-Language-Action Flow Policies},
  author={Reuss, Moritz and Zhou, Hongyi and R{\"u}hle, Marcel and Ya{\u{g}}murlu, {\"O}mer Erdin{\c{c}} and Otto, Fabian and Lioutikov, Rudolf},
  year={2025},
  eprint={2509.04996},
  archivePrefix={arXiv},
  primaryClass={cs.RO},
  url={https://arxiv.org/abs/2509.04996}
}

@inproceedings{shah2023gnm,
  author={Shah, Dhruv and Sridhar, Ajay and Bhorkar, Arjun and Hirose, Noriaki and Levine, Sergey},
  booktitle={2023 IEEE International Conference on Robotics and Automation (ICRA)},
  title={{GNM}: A General Navigation Model to Drive Any Robot},
  year={2023},
  pages={7226-7233},
  doi={10.1109/ICRA48891.2023.10161227}
}

@inproceedings{navid4d2025,
  author={Liu, Haoran and Wan, Weikang and Yu, Xiqian and Li, Minghan and Zhang, Jiazhao and Zhao, Bo and Chen, Zhibo and Wang, Zhongyuan and Zhang, Zhizheng and Wang, He},
  booktitle={2025 IEEE International Conference on Robotics and Automation (ICRA)},
  title={NaVid-4D: Unleashing Spatial Intelligence in Egocentric {RGB-D} Videos for Vision-and-Language Navigation},
  year={2025},
  pages={10607-10615},
  doi={10.1109/ICRA55743.2025.11128467}
}

@misc{AerialVLA2026,
  title={AerialVLA: A Vision-Language-Action Model for {UAV} Navigation via Minimalist End-to-End Control},
  author={Xu, Peng and Deng, Zhengnan and Deng, Jiayan and Gu, Zonghua and Wan, Shaohua},
  year={2026},
  eprint={2603.14363},
  archivePrefix={arXiv},
  primaryClass={cs.CV},
  url={https://arxiv.org/abs/2603.14363}
}

@misc{AutoFly2026,
  title={AutoFly: Vision-Language-Action Model for {UAV} Autonomous Navigation in the Wild},
  author={Sun, Xiaolou and Si, Wufei and Ni, Wenhui and Li, Yuntian and Wu, Dongming and Xie, Fei and Guan, Runwei and Xu, He-Yang and Ding, Henghui and Wu, Yuan and Yue, Yutao and Huang, Yongming and Xiong, Hui},
  year={2026},
  eprint={2602.09657},
  archivePrefix={arXiv},
  primaryClass={cs.RO},
  url={https://arxiv.org/abs/2602.09657}
}

@misc{Adang2025Singer,
  title={{SINGER}: An Onboard Generalist Vision-Language Navigation Policy for Drones},
  author={Adang, Maximilian and Low, JunEn and Shorinwa, Ola and Schwager, Mac},
  year={2025},
  eprint={2509.18610},
  archivePrefix={arXiv},
  primaryClass={cs.RO},
  url={https://arxiv.org/abs/2509.18610}
}

@article{zhang2024vision,
  title={Vision-and-Language Navigation Today and Tomorrow: A Survey in the Era of Foundation Models},
  author={Zhang, Yue and Ma, Ziqiao and Li, Jialu and Qiao, Yanyuan and Wang, Zun and Chai, Joyce and Wu, Qi and Bansal, Mohit and Kordjamshidi, Parisa},
  journal={Transactions on Machine Learning Research},
  issn={2835-8856},
  year={2024},
  url={https://openreview.net/forum?id=yiqeh2ZYUh}
}

@misc{Kim2026Guardrails,
  title={Modular Safety Guardrails Are Necessary for Foundation-Model-Enabled Robots in the Real World},
  author={Kim, Joonkyung and Chen, Wenxi and Soleymanzadeh, Davood and Ding, Yi and Gao, Xiangbo and Tu, Zhengzhong and Zhang, Ruqi and Fei, Fan and Veer, Sushant and Lyu, Yiwei and Zheng, Minghui and Gu, Yan},
  year={2026},
  eprint={2602.04056},
  archivePrefix={arXiv},
  primaryClass={eess.SY},
  url={https://arxiv.org/abs/2602.04056}
}

@inproceedings{werby2024hierarchical,
  title={Hierarchical Open-Vocabulary {3D} Scene Graphs for Language-Grounded Robot Navigation},
  author={Werby, Abdelrhman and Huang, Chenguang and B{\"u}chner, Martin and Valada, Abhinav and Burgard, Wolfram},
  booktitle={First Workshop on Vision-Language Models for Navigation and Manipulation at ICRA 2024},
  year={2024}
}

@misc{wei2025dualvln,
  title={Ground Slow, Move Fast: A Dual-System Foundation Model for Generalizable Vision-and-Language Navigation},
  author={Wei, Meng and Wan, Chenyang and Peng, Jiaqi and Yu, Xiqian and Yang, Yuqiang and Feng, Delin and Cai, Wenzhe and Zhu, Chenming and Wang, Tai and Pang, Jiangmiao and Liu, Xihui},
  year={2025},
  eprint={2512.08186},
  archivePrefix={arXiv},
  primaryClass={cs.RO},
  url={https://arxiv.org/abs/2512.08186}
}

@misc{qwen3.5,
  title  = {{Qwen3.5}: Towards Native Multimodal Agents},
  author = {{Qwen Team}},
  year   = {2026},
  url    = {https://qwen.ai/blog?id=qwen3.5}
}

@article{zhou2020egoplanneresdffreegradientbasedlocal,
  author={Zhou, Xin and Wang, Zhepei and Ye, Hongkai and Xu, Chao and Gao, Fei},
  journal={IEEE Robotics and Automation Letters},
  title={{EGO-Planner}: An {ESDF}-Free Gradient-Based Local Planner for Quadrotors},
  year={2021},
  volume={6},
  number={2},
  pages={478-485},
  doi={10.1109/LRA.2020.3047728}
}

@article{eschmann2026raptorfoundationpolicyquadrotor,
  author = {Eschmann, Jonas and Albani, Dario and Loianno, Giuseppe},
  title = {{RAPTOR}: A Foundation Policy for Quadrotor Control},
  journal = {Science Robotics},
  volume = {11},
  number = {114},
  pages = {eaec1481},
  year = {2026},
  doi = {10.1126/scirobotics.aec1481}
}

@inproceedings{liu2023aerialvln,
  title={AerialVLN: Vision-and-Language Navigation for {UAV}s},
  author={Liu, Shubo and Zhang, Hongsheng and Qi, Yuankai and Wang, Peng and Zhang, Yanning and Wu, Qi},
  booktitle={Proceedings of the IEEE/CVF International Conference on Computer Vision},
  pages={15384--15394},
  year={2023}
}

@misc{wu2025vlaan,
  title={{VLA-AN}: An Efficient and Onboard Vision-Language-Action Framework for Aerial Navigation in Complex Environments},
  author={Wu, Yuze and Zhu, Mo and Li, Xingxing and Du, Yuheng and Fan, Yuxin and Li, Wenjun and Han, Zhichao and Zhou, Xin and Gao, Fei},
  year={2025},
  eprint={2512.15258},
  archivePrefix={arXiv},
  primaryClass={cs.RO},
  url={https://arxiv.org/abs/2512.15258}
}

@inproceedings{hu2025spf,
  title = {See, Point, Fly: A Learning-Free {VLM} Framework for Universal Unmanned Aerial Navigation},
  author = {Hu, Chih Yao and Lin, Yang-Sen and Lee, Yuna and Su, Chih-Hai and Lee, Jie-Ying and Tsai, Shr-Ruei and Lin, Chin-Yang and Chen, Kuan-Wen and Ke, Tsung-Wei and Liu, Yu-Lun},
  booktitle = {Proceedings of The 9th Conference on Robot Learning},
  pages = {4697--4708},
  year = {2025},
  volume = {305},
  series = {Proceedings of Machine Learning Research},
  publisher = {PMLR}
}

@misc{xu2026fly0,
  title={Fly0: Decoupling Semantic Grounding from Geometric Planning for Zero-Shot Aerial Navigation},
  author={Xu, Zhenxing and Lu, Brikit and Bao, Weidong and Zhu, Zhengqiu and Zhang, Junsong and Yan, Hui and Lu, Wenhao and Wang, Ji},
  year={2026},
  eprint={2602.15875},
  archivePrefix={arXiv},
  primaryClass={cs.RO},
  url={https://arxiv.org/abs/2602.15875}
}

@misc{chen2026airhunt,
  title={AirHunt: Bridging {VLM} Semantics and Continuous Planning for Efficient Aerial Object Navigation},
  author={Chen, Xuecheng and Liu, Zongzhuo and Ma, Jianfa and Du, Bang and Zhang, Tiantian and Wang, Xueqian and Zhou, Boyu},
  year={2026},
  eprint={2601.12742},
  archivePrefix={arXiv},
  primaryClass={cs.RO},
  url={https://arxiv.org/abs/2601.12742}
}

@misc{zheng2026onfly,
  title={OnFly: Onboard Zero-Shot Aerial Vision-Language Navigation toward Safety and Efficiency},
  author={Zheng, Guiyong and Ban, Yueting and Zhang, Mingjie and Zheng, Juepeng and Zhou, Boyu},
  year={2026},
  eprint={2603.10682},
  archivePrefix={arXiv},
  primaryClass={cs.RO},
  url={https://arxiv.org/abs/2603.10682}
}

@article{kaufmann2023champion,
  title={Champion-Level Drone Racing Using Deep Reinforcement Learning},
  author={Kaufmann, Elia and Bauersfeld, Leonard and Loquercio, Antonio and M{\"u}ller, Matthias and Koltun, Vladlen and Scaramuzza, Davide},
  journal={Nature},
  volume={620},
  number={7976},
  pages={982--987},
  year={2023},
  publisher={Nature Publishing Group}
}

@inproceedings{LARS,
  author={Tayar, Marco S. and de Oliveira, Lucas K. and Tommaselli, Felipe Andrade G. and Negri, Juliano D. and Segreto, Thiago H. and Godoy, Ricardo V. and Becker, Marcelo},
  booktitle={2025 Latin American Robotics Symposium (LARS)},
  title={Autonomous {UAV} Flight Navigation in Confined Spaces: A Reinforcement Learning Approach},
  year={2025},
  pages={1-6},
  doi={10.1109/LARS69345.2025.11273007}
}

@misc{deoliveira2026mira,
      title={MIRA: A Modular Open-Source Micro-UAV for Indoor Research}, 
      author={Lucas K. de Oliveira and Felipe A. G. Tommaselli and João Aires Marsicano and Marco S. Tayar and Pedro A. R. Saraiva and Ricardo V. Godoy and Marcelo Becker},
      year={2026},
      eprint={2607.11785},
      archivePrefix={arXiv},
      primaryClass={cs.RO},
}

\end{document}